\documentclass[journal]{IEEEtran}
\usepackage{latexsym,amssymb,amsmath,graphicx,epstopdf,epsf,epsfig,cite,bbm,float,subfig,graphics}

\def\ninept{\def\baselinestretch{1}}
\ninept

\newcommand{\sy}{\left\{y_{t}\right\}_{t \geq 1}}

\newcommand{\be}{\begin{equation}}
\newcommand{\ee}{\end{equation}}
\newcommand{\bea}{\begin{eqnarray}}
\newcommand{\eea}{\end{eqnarray}}
\newcommand{\nn}{\nonumber}

\newcommand{\MB}{\left[\begin{array}}
\newcommand{\ME}{\end{array}\right]}

\renewcommand{\vec}[1]{\mbox{\boldmath${#1}$}}

\newcommand{\vu}{\vec{u}}
\newcommand{\vw}{\vec{w}}
\newcommand{\ei}{\end{itemize}}
\newcommand{\bi}{\begin{itemize}}

\newcommand{\vx}{\mbox{$\vec{x}$}}

\newcommand{\lp}{\lambda^+}

\newcommand{\defi}{\stackrel{\bigtriangleup}{=}}

\newcommand{\eps}{\mbox{$\epsilon$}}
\begin{document}
\title{A Deterministic Analysis of an Online Convex Mixture of Expert Algorithms}
\author{Mehmet~A.~Donmez, Sait Tunc and Suleyman~S.~Kozat,~\IEEEmembership{Senior Member}
\thanks{ This work is supported in part by IBM Faculty Award and Outstanding Young Scientist Award Program, Turkish Academy of Sciences. Suleyman S. Kozat, Mehmet A. Donmez and Sait Tunc (\{skozat,medonmez,saittunc\}@ku.edu.tr) are with the Competitive Signal Processing Laboratory at Koc University, Istanbul, tel: +902123381864.}}

\maketitle

\begin{abstract}
We analyze an online learning algorithm that adaptively combines outputs of two constituent
algorithms (or the experts) running in parallel to model an unknown desired signal. This online learning algorithm
is shown to achieve (and in some cases outperform) the mean-square error (MSE) performance of
the best constituent algorithm in the mixture in
the steady-state. However, the MSE analysis of this algorithm in the literature
uses approximations and relies on statistical models on the underlying signals and
systems. Hence, such an analysis may not be useful or valid for
signals generated by various real life systems that show high degrees
of nonstationarity, limit cycles and, in many cases, that are even chaotic.
In this paper, we produce results in an individual sequence manner.
In particular, we relate the time-accumulated squared estimation error of this online
algorithm at any time over any interval to the time-accumulated
squared estimation error of the optimal convex mixture of the
constituent algorithms directly tuned to the underlying signal
in a deterministic sense without any statistical assumptions. In
this sense, our analysis provides the transient, steady-state and
tracking behavior of this algorithm in a "strong" sense without
any approximations in the derivations or statistical assumptions
on the underlying signals such that our results are guaranteed
to hold. We illustrate the introduced results through examples.
\end{abstract}

\IEEEpeerreviewmaketitle

\section{Introduction}
\label{sec:introduction}
The problem of estimating or learning an unknown desired signal is heavily
investigated in online learning \cite{NNLS1,NNLS2,NNLS3,NNLS4,cesabook,KiWa02,cesab}
and adaptive signal processing literature \cite{sinfed,convex,kozat,sayed}.
However, in various applications, certain difficulties arise in the estimation
process due to the lack of structural and statistical information
about the data model. To resolve this lack of information, mixture
approaches are proposed that adaptively combine outputs of multiple
constituent algorithms performing the same task in the online learning literature under the mixture of
experts framework \cite{KiWa02,cesab,cesabook} and adaptive signal processing
under the adaptive mixture methods framework \cite{sinfed,convex,kozat}. These parallel running algorithms can be seen as
alternative hypotheses for modeling, which can be exploited for both
performance improvement and robustness. Along these lines, an online convexly
constrained mixture method that combines outputs of two learning algorithms
is introduced in \cite{convex}. In this approach, the outputs
of the constituent algorithms that run in parallel on the same task are adaptively combined under a convex
constraint to minimize the final MSE. This adaptive mixture is shown
to be universal with respect to the input algorithms in a certain
stochastic sense such that this mixture achieves (and in some cases outperforms)
the MSE performance of the best constituent algorithm in the mixture in
the steady-state \cite{convex}. However, the MSE analysis of this adaptive mixture
for the steady-state and during the transient regions uses
approximations, e.g., separation assumptions, and relies on
statistical models on the signals and systems, e.g., stationary
data models \cite{convex,kozat}. In this paper, we study this algorithm
from the perspective of online learning and produce results in an individual
sequence manner such that our results are guaranteed to hold for
any bounded arbitrary signal.

Nevertheless, signals produced by various real life systems, such as
in underwater acoustic communication applications, show high degrees
of nonstationarity, limit cycles and, in many cases, are even chaotic
so that they hardly fit to assumed statistical models \cite{kozatPhd}. Hence an analysis
based on certain statistical assumptions or approximations may not
be useful or adequate under these conditions. To this end, we refrain
from making any statistical assumptions on the underlying signals and
present an analysis that is guaranteed to hold for any bounded
arbitrary signal without any approximations.
In particular, we relate the performance of this learning algorithm that
adaptively combines outputs of two constituent algorithms to the performance
of the optimal convex combination that is directly tuned to the underlying
signal and outputs of the constituent algorithms in a deterministic
sense. Naturally, this optimal convex combination can only be chosen
in hindsight after observing the whole signal and outputs a
priori (before we even start processing the data). Since we compare
the performance of this algorithm with respect to the best
convex combination of the constituent filters in a deterministic
sense over any time interval, our analysis provides, without any
assumptions, the transient, the tracking and the steady-state
behaviors together \cite{KiWa02,cesab,cesabook}. In particular, if the analysis window
starts from $t = 1$, then we obtain the transient behavior; if the
window length goes to infinity, then we obtain the steadystate
behavior; and finally if the analyze window is selected
arbitrary, then we get the tracking behavior as explained in
detail in Section III. The corresponding bounds may also hold
for unbounded signals such as with Gaussian and Laplacian
distributions, if one can define reasonable bounds such that
the effect of samples of the desired signal that are outside of
an interval on the cumulative loss diminishes as the data size
increases as demonstrated in Section III.

After we provide a brief system description in
Section~\ref{sec:problem_description}, we present a deterministic
analysis of the convexly constrained mixture algorithm in
Section~\ref{sec:deterministic_analysis}, where the performance bounds
are given as a theorem and a lemma. We illustrate the introduced results
through examples in Section~\ref{sec:examples}. The paper concludes with certain
remarks.

\section{Problem Description}\label{sec:problem_description}
In this framework, we have a desired signal $\sy$, where $|y_{t}| \leq
Y <\infty$, and two constituent algorithms running in parallel producing
$\{\hat{y}_{1,t}\}_{t \geq 1}$ and $\{\hat{y}_{2,t}\}_{t \geq 1}$,
respectively, as the estimations (or predictions) of the desired
signal $\sy$. We assume that $Y$ is known.
Here, we have no restrictions on $\hat{y}_{1,t}$
or $\hat{y}_{2,t}$, e.g., these outputs
are not required to be causal, however, without loss of generality, we
assume $|\hat{y}_{1,t}| \leq Y$ and $|\hat{y}_{2,t}| \leq Y$, i.e.,
these outputs can be clipped to the range $[-Y,Y]$ without sacrificing
performance under the squared error. As an example, the desired signal
and outputs of the constituent learning algorithms can be single realizations
generated under the framework of \cite{convex}. At each time $t$, the
convexly constrained algorithm receives an input vector $\vx_{t} \defi
[\hat{y}_{1,t}\;\hat{y}_{2,t}]^T$ and outputs
\begin{align*}
\hat{y}_{t} &= \lambda_{t}  \hat{y}_{1,t} + (1-\lambda_{t}) \hat{y}_{2,t}= \vw_t^T \vx_{t},
\end{align*}\normalsize
where $\vw_t\defi[\lambda_{t} \; (1-\lambda_{t})]^T$, $0 \leq \lambda_{t} \leq 1$, as the final estimate. The final estimation error is given by $e_{t}=y_{t} -\hat{y}_{t}$.

The combination weight $\lambda_{t}$ is trained through an
auxiliary variable using a stochastic gradient update to minimize the
squared final estimation error as
\begin{align}
& \lambda_{t}  = \frac{1}{1+e^{-\rho_{t}}} \label{eq:son2}, \\
& \rho_{t+1}  = \rho_{t}-\mu \nabla_{\rho}e^2_{t}\big|_{\rho=\rho_{t}} \nonumber \\
  & =  \rho_{t}+ \mu e_{t}\lambda_{t}(1-\lambda_{t}) [\hat{y}_{1,t}-\hat{y}_{2,t}], \label{eq:1}
\end{align}\normalsize
where $\mu > 0$ is the learning rate. The combination
parameter $\lambda_{t}$ in \eqref{eq:son2} is constrained to lie in
$[\lambda^+,(1-\lambda^+)]$, $0<\lambda^+ < 1/2$ in \cite{convex},
since the update in \eqref{eq:1} may slow down when $\lambda_{t}$ is
too close to the boundaries. We follow the same restriction and
analyze \eqref{eq:1} under this constraint. The algorithm is presented in Table~\ref{table:alg}.

\begin{table}
\begin{tabular}[t]{|l|}
\hline
{\bf The Convexly Constrained Algorithm:} \\
\hline
\hspace*{0.1in}{\bf Parameters:}\\
\hspace*{0.2in}$\mu>0$: learning rate.\\
\hspace*{0.1in}{\bf Inputs:}\\
\hspace*{0.2in}$y_t$: desired signal. \\
\hspace*{0.2in}$\hat{y}_{1,t},\hat{y}_{2,t}$: constituent learning algorithms. \\
\hspace*{0.1in}{\bf Outputs:}\\
\hspace*{0.2in}$\hat{y}_{t}$: estimate of the desired signal.\\
\hline
\hspace*{0.1in}{\bf Initialization:} Set the initial weights $\lambda_1=1/2$ and $\rho_1=0$. \\
\hspace*{0.1in}for $t=1:\ldots:n$, \\
\hspace*{0.2in}$\%$ receive the constituent algorithm outputs $\hat{y}_{1,t}$ and $\hat{y}_{2,t}$ and\\
\hspace*{0.2in}$\%$ estimate the desired signal\\
\hspace*{0.2in}$\hat{y}_t=\lambda_t \hat{y}_{1,t} + (1-\lambda_t)\hat{y}_{2,t}$  \\
\hspace*{0.2in}$\%$ Upon receiving $y_t$, update the weight according to the rule:\\
\hspace*{0.2in}$\rho_{t+1} =  \rho_{t}+ \mu e_{t}\lambda_{t}(1-\lambda_{t}) [\hat{y}_{1,t}-\hat{y}_{2,t}]$  \\
\hspace*{0.2in}$\lambda_{t+1}  = \frac{1}{1+e^{-\rho_{t+1}}}$\\
\hspace*{0.1in}{endfor} \\
\hline
\end{tabular}
\caption{The learning algorithm that adaptively combines outputs of two algorithms.}
\label{table:alg}
\end{table}

Under the deterministic analysis framework, the performance of the algorithm is determined
by the time-accumulated squared error \cite{cesa98,cesab,vovk,war,cesabook}.
When applied to any sequence $\sy$, the algorithm of \eqref{eq:son2}
yields the total accumulated loss
\begin{align}
L_n(\hat{y},y) =L_n(\vw_t^T\vx_t,y)\defi \sum_{t=1}^n (y_{t}-\hat{y}_{t})^2\label{eq:cumulative}
\end{align}
for any $n$.
We emphasize that for unbounded signals such as Gaussian and Laplacian distributions, we can define
a suitable $Y$ such that the samples of $y_t$ are inside of the interval $[-Y,Y]$ with
high probability and the effect of the samples that are outside of this interval on
the cumulative loss \eqref{eq:cumulative} diminishes as $n$ gets larger.

We next provide deterministic bounds on $L_n(\hat{y},y)$ with respect to the best convex combination $
\min\limits_{\beta\in[0,1]} L_n(\hat{y}_{\beta},y)$, where
\[
 L_n(\hat{y}_{\beta},y) = L_n(\vu^T\vx_t,y)=\sum_{t=1}^n (y_{t}-\hat{y}_{\beta,t})^2
\]
and
\begin{align*}
\hat{y}_{\beta,t}&\defi\beta \hat{y}_{1,t}+(1-\beta)\hat{y}_{2,t}= \vu^T \vx_{t},
\end{align*}
$\vu \defi [\beta\;1-\beta]^T$, that holds uniformly in an individual sequence manner without any
stochastic assumptions on $y_{t}$, $\hat{y}_{1,t}$, $\hat{y}_{2,t}$ or
$n$.
Note that the best fixed convex combination parameter
\begin{align*}
\beta_o = \arg \min\limits_{\beta\in[0,1]} L_n(\hat{y}_{\beta},y)
\end{align*} and the corresponding estimator
\begin{align*}
\hat{y}_{\beta_o,t}=\beta_o \hat{y}_{1,t}+(1-\beta_o)\hat{y}_{2,t},
\end{align*}
which we compare
the performance against,
can only be determined after observing the entire sequences, i.e.,
$\{y_{t}\},\{\hat{y}_{1,t}\}$ and $\{\hat{y}_{2,t}\}$, in advance for all $n$.

\section{A Deterministic Analysis \label{sec:deterministic_analysis}}

In this section, we first relate the accumulated loss of the mixture
to the accumulated loss of the best convex combination that
minimizes the accumulated loss in the following theorem. Then, we
demonstrate that one cannot improve the convergence rate of this upper
bound using our methodology directly and the Kullback-Leibler (KL)
divergence \cite{KiWa02} as the distance measure by providing counter
examples as a lemma. The use of the KL divergence as a distance measure
for obtaining worst-case loss bounds was pioneered by Littlestone \cite{littlestone},
and later adopted extensively in the online learning literature \cite{CeLoWa:96,KiWa02,cesab}.
We emphasize that although the steady-state and
transient MSE performances of the convexly constrained mixture
algorithm are analyzed with respect to the constituent learning algorithms
\cite{convex,kozat}, we perform the steady-state, transient and tracking analysis without any stochastic assumptions or use any
approximations in the following theorem.\\

\noindent
{\bf Theorem:} The algorithm given in \eqref{eq:1}, when applied to
any sequence $\sy$, with $|y_{t}| \leq Y<\infty$, yields, for any $n$ and $\epsilon>0$
\begin{equation}
L_n(\hat{y},y)-\left( \frac{2 \epsilon+1}{1-z^2}\right) \min\limits_{\beta\in[0,1]} \left\{ L_n(\hat{y}_{\beta},y)\right\} \leq O\left( \frac{1}{\epsilon} \right), \label{eq:theorem}
\end{equation}\normalsize
where $O\left(.\right)$ is the order notation, $\hat{y}_{\beta,t}=\beta \hat{y}_{1,t}+(1-\beta)\hat{y}_{2,t}$, $z\defi \frac{1-4 \lambda^+(1-\lambda^+)}{1+4
  \lambda^+(1-\lambda^+)} < 1$ and step size $\mu = \frac{4 \epsilon}{2\epsilon+1}\frac{2+2z}{Y^2}$, provided that $\lambda_{t}
\in \left[\lambda^+,1-\lambda^+\right]$ for all
$t$ during the adaptation. \\

This theorem provides a regret bound for the algorithm \eqref{eq:1} showing that the cumulative loss of the convexly constrained algorithm
is close to a factor times the cumulative loss of the algorithm with the best weight chosen in hindsight. If we define the regret
\begin{equation}
R_n \defi L_n(\hat{y},y)-\left( \frac{2 \epsilon+1}{1-z^2}\right) \min\limits_{\beta\in[0,1]} \left\{ L_n(\hat{y}_{\beta},y)\right\},\label{eq:regret}
\end{equation}
then equation \eqref{eq:theorem} implies that time-normalized regret
\begin{align*}
\frac{R_n}{n} \defi \frac{L_n(\hat{y},y)}{n}-\left( \frac{2 \epsilon+1}{1-z^2}\right) \min\limits_{\beta\in[0,1]} \left\{ \frac{L_n(\hat{y}_{\beta},y)}{n}\right\}
\end{align*}
converges to zero at a rate $O\left( \frac{1}{n\epsilon} \right)$ uniformly
over the desired signal and the outputs of constituent algorithms.
Moreover, \eqref{eq:theorem} provides the exact trade-off between the
transient and steady-state performances of the convex mixture in a
deterministic sense without any assumptions or approximations.
Note that \eqref{eq:theorem} is guaranteed to hold independent of the
initial condition of the combination weight $\lambda_t$
for any time interval in an individual sequence manner.
Hence, \eqref{eq:theorem} also provides the tracking performance of
the convexly constrained algorithm in a deterministic sense.
From \eqref{eq:theorem},
we observe that the convergence rate of the right hand
side, i.e., the bound, is $O\left( \frac{1}{n\epsilon} \right)$, and, as in the
stochastic case \cite{kozat}, to get a tighter asymptotic bound with
respect to the optimal convex combination of the learning algorithms, we require a
smaller $\epsilon$, i.e., smaller learning rate $\mu$, which increases
the right hand side of \eqref{eq:theorem}.  Although this result is
well-known in the adaptive filtering literature and appears widely in
stochastic contexts, however, this trade-off is guaranteed to hold in
here without any statistical assumptions or approximations.  Note that
the optimal convex combination in \eqref{eq:theorem}, i.e., minimizing $\beta$,
depends on the entire signal and outputs of the
constituent algorithms for all $n$ and hence it can only be determined in hindsight.\\

\noindent
{\bf Proof:} To prove the theorem, we use the approach introduced in
\cite{cesab} (and later used in \cite{KiWa02}) based on
measuring progress of a mixture algorithm using certain distance
measures.

We first convert \eqref{eq:1} to a direct update on $\lambda_{t}$ and
use this direct update in the proof. Using
\begin{align*}
e^{-\rho_{t}} = \frac{1-\lambda_{t}}{\lambda_{t}}
\end{align*}
from \eqref{eq:son2}, the update in \eqref{eq:1} can be written as
\begin{align}
\lambda_{t+1} &= \frac{1}{1+e^{-\rho_{t+1}}}\nonumber\\
             & =  \frac{1}{1+e^{-\rho_{t}- \mu e_{t}\lambda_{t}(1-\lambda_{t}) [\hat{y}_{1,t}-\hat{y}_{2,t}]}} \nonumber \\
             & = \frac{1}{1+ \frac{1-\lambda_{t}}{\lambda_{t}} e^{-\mu e_{t}\lambda_{t}(1-\lambda_{t}) [\hat{y}_{1,t}-\hat{y}_{2,t}]}} \nonumber \\
             & = \frac{\lambda_{t}e^{\mu e_{t} \lambda_{t} (1-\lambda_{t}) \hat{y}_{1,t}}}{\lambda_{t}e^{\mu e_{t} \lambda_{t} (1-\lambda_{t}) \hat{y}_{1,t}} + (1-\lambda_{t})e^{\mu e_{t} \lambda_{t} (1-\lambda_{t}) \hat{y}_{2,t}}}.
\label{update}
\end{align}\normalsize
Unlike \cite{KiWa02} (Lemma 5.8), our update in \eqref{update} has,
in a certain sense, an adaptive learning rate $\mu \lambda_{t}
(1-\lambda_{t})$ which requires different formulation, however, follows
similar lines of \cite{KiWa02} in certain parts.

Here, for a fixed $\beta\in[0,1]$, we define an estimator
\begin{align*}
\hat{y}_{\beta,t}&\defi\beta \hat{y}_{1,t}+(1-\beta)\hat{y}_{2,t}=\vu^T \vx_{t},
\end{align*}
where $\beta\in[0,1]$ and $\vu \defi [\beta\;\;1-\beta]^T$. Defining
\begin{align*}
\zeta_{t} = e^{\mu e_{t}\lambda_{t}(1 - \lambda_{t})},
\end{align*}
we have from \eqref{update}
\begin{align}
&\beta \ln\left(\frac{\lambda_{t+1}}{\lambda_{t}}\right)+(1-\beta) \ln\left(\frac{1-\lambda_{t+1}}{1-\lambda_{t}}\right) \nn\\&= \hat{y}_{\beta,t} \ln \zeta_{t} - \ln\left(\lambda_{t} \zeta_{t}^{\hat{y}_{1,t}} + (1-\lambda_{t})\zeta_{t}^{\hat{y}_{2,t}}\right).\label{eq:b}
\end{align}\normalsize
Using the inequality
\begin{align*}
\alpha^x \leq 1 - x(1-\alpha)
\end{align*} for $\alpha \geq 0$ and $x \in [0,1]$ from \cite{cesab}, we have
\begin{align*}
\zeta_{t}^{\hat{y}_{1,t}} & = (\zeta_{t}^{2Y})^{\frac{\hat{y}_{1,t} + Y}{2Y}} \zeta_{t}^{-Y} \\
                   & \leq \zeta_{t}^{-Y}\left(1 - \frac{\hat{y}_{1,t} + Y}{2Y}(1- \zeta_{t}^{2Y})\right), \nonumber
\end{align*}\normalsize
which implies in \eqref{eq:b}
\begin{align}
&\ln\left(\lambda_t \zeta_{t}^{\hat{y}_{1,t}} + (1-\lambda_t) \zeta_{t}^{\hat{y}_{2,t}}\right) \nn\\
&\leq \ln \left( \zeta_{t}^{-Y}(1 - \frac{\lambda \hat{y}_{1,t} + (1-\lambda_t) \hat{y}_{2,t} + Y}{2Y}(1- \zeta_{t}^{2Y})) \right) \nonumber \\
& = -Y \ln \zeta_{t} + \ln \left(1 - \frac{\hat{y}_{t}+ Y}{2Y}(1-\zeta_{t}^{2Y})\right),\label{eq:ln}
\end{align}\normalsize
where $\hat{y}_{t}= \lambda_{t} \hat{y}_{1,t} + (1-\lambda_{t})
\hat{y}_{2,t}$. As in \cite{KiWa02}, one can further bound
\eqref{eq:ln} using
\begin{align*}
\ln\left(1-q(1-e^p)\right) \leq pq+\frac{p^2}{8}
\end{align*}
for $0\leq q<1$ (originally from \cite{cesab})
\begin{align}
&\ln\left(\lambda_t \zeta_{t}^{\hat{y}_{1,t}} + (1-\lambda_t) \zeta_{t}^{\hat{y}_{2,t}}\right) \nn\\
&\leq  -Y \ln \zeta_{t} +  (\hat{y}_{t}+ Y) \ln\zeta_{t} +\frac{Y^2 (\ln \zeta_{t})^2}{2}. \label{eq:a}
\end{align}\normalsize

Using
\eqref{eq:a} in \eqref{eq:b} yields
\begin{align}
&\beta \ln\left(\frac{\lambda_{t+1}}{\lambda_{t}}\right)+(1-\beta) \ln\left(\frac{1-\lambda_{t+1}}{1-\lambda_{t}}\right) \geq \label{eq:c}\\ & (\hat{y}_{\beta,t} + Y)\ln \zeta_{t} -  (\hat{y}_{t}+ Y) \ln\zeta_{t} - \frac{Y^2 (\ln \zeta_{t})^2}{2} . \nn
\end{align}\normalsize

At each adaptation, the progress made by the algorithm towards $\vu$ at time
$t$ is measured as $D(\vu||\vw_{t}) - D(\vu||\vw_{t+1})$, where $\vw_{t}\defi [\lambda_{t} \; (1-\lambda_{t})]^T$ and
\begin{align*}
D(\vu||\vw) \defi \sum_{i=1}^2 u_i \ln (u_i/w_i)
\end{align*}
is the KL divergence \cite{cesab,cover}, $\vu \in \left[0,1\right]^2$, $\vw \in \left[0,1\right]^2$. We require that this
progress is at least $a(y_{t}-\hat{y}_{t})^2 - b(y_{t}-\hat{y}_{\beta,t})^2$
for certain $a$, $b$, $\mu$ \cite{cesab,KiWa02}, i.e.,
\begin{align}
&a(y_{t}-\hat{y}_{t})^2 - b(y_{t}-\hat{y}_{\beta,t})^2 \nn\\
&\leq D(\vu||\vw_{t}) - D(\vu||\vw_{t+1}) \nn\\
&= \beta \ln\left(\frac{\lambda_{t+1}}{\lambda_{t}}\right)+(1-\beta) \ln\left(\frac{1-\lambda_{t+1}}{1-\lambda_{t}}\right),
\label{desired_bound}
\end{align}
which yields the desired deterministic bound in \eqref{eq:theorem}
after telescoping.

In information theory and probability theory, the KL divergence, which is also known as the
relative entropy, is
empirically shown to be an efficient measure of the distance between two probability vectors \cite{KiWa02,cesab,cover}.
Here, the vectors $\vu$ and $\vw_{t}$ are probability vectors, i.e., $\vu,\vw_t\in[0,1]^2$ and $\vu^T\vec{1}=\vw_t^T\vec{1}=1$, where $\vec{1}\defi[1\;1]^T$.
This use of KL divergence as a distance measure between weight vectors is widespread in the online learning literature \cite{KiWa02,cesa98,CeLoWa:96}.

We observe from \eqref{desired_bound} and \eqref{eq:c} that to prove the
theorem, it is sufficient to show that $G(y_{t},\hat{y}_{t},\hat{y}_{\beta,t},\zeta_{t})\leq 0$, where
\begin{align}
&G(y_{t},\hat{y}_{t},\hat{y}_{\beta,t},\zeta_{t}) \defi -(\hat{y}_{\beta,t} + Y)\ln \zeta_{t} +  (\hat{y}_{t}+ Y) \ln\zeta_{t}  \nn \\ & +\frac{Y^2 (\ln \zeta_{t})^2}{2} + a(y_{t}-\hat{y}_{t})^2 - b(y_{t}-\hat{y}_{\beta,t})^2.\label{main_func}
\end{align}
For fixed $y_{t},\hat{y}_{t},\zeta_{t}$,
$G(y_{t},\hat{y}_{t},\hat{y}_{\beta,t},\zeta_{t})$ is maximized when
$\frac{\partial G}{\partial {\hat{y}_{\beta,t}}}=0$, i.e.,
\begin{equation*}
\hat{y}_{\beta,t} -
y_{t}+ \frac{\ln \zeta_{t}}{2b} = 0
\end{equation*}
since $\frac{\partial^2 G}{\partial {\hat{y}_{\beta,t}}^2} = -2b < 0$,
yielding
$
{\hat{y}_{\beta,t}}^* = y_{t}- \frac{\ln \zeta_{t}}{2b}
$.
Note that while taking the partial derivative of $G(\cdot)$ with
respect to $\hat{y}_{\beta,t}$ and finding ${\hat{y}_{\beta,t}}^*$, we assume
that all $y_{t},\hat{y}_{t},\zeta_{t}$ are fixed, i.e., their partial
derivatives with respect to $\hat{y}_{\beta,t}$ is zero. This yields an
upper bound on $G(\cdot)$ in terms of $\hat{y}_{\beta,t}$. Hence, it is
sufficient to show that
$
G(y_{t},\hat{y}_{t},{\hat{y}_{\beta,t}}^*,\zeta_{t})\leq 0
$
such that \cite{KiWa02}
\begin{align}
& G(y_{t},\hat{y},{\hat{y}_{\beta,t}}^*,\zeta_{t}) \nonumber \\
& = - \left(y_{t}+ Y - \frac{\ln \zeta_{t}}{2b}\right) \ln \zeta_{t} +   (\hat{y}_{t}+ Y) \ln\zeta_{t} \nonumber \\
& +\frac{Y^2 (\ln \zeta_{t})^2}{2}  + a(y_{t}-\hat{y}_{t})^2 - \frac{(\ln \zeta_{t} )^2}{4b}\label{eq:ust1} \\
& = a (y_{t}-\hat{y}_{t})^2 - (y_{t}-\hat{y}_{t})\ln \zeta_{t} + \frac{(\ln \zeta_{t})^2}{4b} \nn \\
&+ \frac{Y^2 (\ln \zeta_{t})^2}{2}\nonumber\\
& = (y_{t} - \hat{y}_{t})^2\times \Bigg[  a - \mu\lambda_{t}(1-\lambda_{t})   \nn \\
&+\frac{{\mu}^2{\lambda_{t}}^2 (1-\lambda_{t})^2}{4b} + \frac{Y^2 {\mu}^2 {\lambda_{t}}^2 (1-\lambda_{t})^2}{2} \Bigg]. \label{eq:last}
\end{align}\normalsize

For \eqref{eq:last} to be negative, defining $k \defi \lambda_{t} (1-\lambda_{t})$ and
\[
H(k) \defi k^2 \mu^2 (\frac{Y^2}{2} + \frac{1}{4b}) -
\mu k + a,
\]\normalsize
it is sufficient to show that $H(k) \leq 0$ for $k \in [\lp (1-\lp) ,
  \frac{1}{4}]$, i.e., $ k \in [\lp (1-\lp) , \frac{1}{4}]$ when
$\lambda_{t} \in [\lp,(1-\lp)]$, since $H(k)$ is a convex quadratic
function of $k$, i.e., $\frac{\partial^2 H}{\partial k^2} > 0$. Hence,
we require the interval where the function $H(\cdot)$ is negative
should include $[\lp(1-\lp),\frac{1}{4}]$, i.e., the roots $k_1$ and
$k_2$ (where $k_2 \leq k_1$) of $H(\cdot)$ should satisfy
\begin{align*}
k_1 &\geq \frac{1}{4},\;k_2 \leq \lp(1-\lp),
\end{align*}
where
\begin{align}
k_{1} & = \frac{\mu + \sqrt{{\mu}^2 - 4 {\mu}^2 a \left(\frac{Y^2}{2} + \frac{1}{4b}\right)}}{2{\mu}^2 (\frac{Y^2}{2} + \frac{1}{4b})} = \frac{1 + \sqrt{1 - 4  a s}}{2\mu s}\label{eq:root1},\\
k_{2} & = \frac{\mu - \sqrt{{\mu}^2 - 4 {\mu}^2 a \left(\frac{Y^2}{2} + \frac{1}{4b}\right)}}{2{\mu}^2 (\frac{Y^2}{2} + \frac{1}{4b})} = \frac{1 - \sqrt{1 - 4  a s}}{2\mu s}\label{eq:root2}
\end{align}
and \[s \defi \left(\frac{Y^2}{2} + \frac{1}{4b}\right).\]

To satisfy $k_1 \geq 1/4$, we straightforwardly require from \eqref{eq:root1}
\[
\frac{2+2 \sqrt{1-4as}}{s} \geq
\mu.
\]\normalsize
To get the tightest upper bound for \eqref{eq:root1}, we set
\[
\mu = \frac{2+2 \sqrt{1-4as}}{s},
\]\normalsize
i.e., the largest allowable learning rate.

To have $k_2 \leq
\lambda^+(1-\lambda^+)$ with $\mu = \frac{2+2 \sqrt{1-4as}}{s}$, from
\eqref{eq:root2} we require \be
\frac{1-\sqrt{1-4as}}{4(1+\sqrt{1-4as})} \leq
\lambda^+(1-\lambda^+). \label{eq:yeter} \ee Equation \eqref{eq:yeter}
yields
\begin{align}
as &= a \left(\frac{Y^2}{2} + \frac{1}{4b}\right) \leq \frac{1- z^2}{4}, \label{eq:3}
\end{align}
where
\[
z \defi \frac{1-4 \lambda^+(1-\lambda^+)}{1+4
  \lambda^+(1-\lambda^+)}
\]\normalsize
and $z < 1$ after some algebra.

To satisfy \eqref{eq:3}, we set $b = \frac{\epsilon}{Y^2}$ for any (or
arbitrarily small) $\epsilon > 0$ that results
\begin{equation}
a \leq \frac{(1-z^2) \epsilon}{Y^2 (2\epsilon+1)}. \label{eq:4}
\end{equation}\normalsize
To get the tightest bound in \eqref{desired_bound}, we select
\begin{equation*}
a = \frac{(1-z^2) \epsilon}{Y^2 (2\epsilon+1)}
\end{equation*} in \eqref{eq:4}. Such selection of $a$, $b$ and $\mu$ results in \eqref{desired_bound}
\begin{align}
&\left(\frac{(1-z^2) \epsilon}{Y^2 (2\epsilon+1)}\right)
(y_{t}-\hat{y}_{t})^2 - \left( \frac{\epsilon}{Y^2}\right)
(y_{t}- \hat{y}_{\beta,t})^2  \nn\\&\leq
\beta \ln\left(\frac{\lambda_{t+1}}{\lambda_{t}}\right)+(1-\beta) \ln\left(\frac{1-\lambda_{t+1}}{1-\lambda_{t}}\right). \label{eq:fin1}
\end{align}\normalsize
After telescoping, i.e., summation over $t$, $\sum_{t=1}^n$,
\eqref{eq:fin1} yields
\begin{align}
&aL_n(\hat{y},y)-b \min\limits_{\beta\in[0,1]} \left\{ L_n(\hat{y}_{\beta},y)\right\} \nn\\
&\leq \beta \ln\left(\frac{\lambda_{n+1}}{\lambda_1}\right)+(1-\beta) \ln\left(\frac{1-\lambda_{n+1}}{1-\lambda_1}\right)\leq  O(1),
\end{align}
so that
\begin{align}
&\left(\frac{(1-z^2) \epsilon}{Y^2 (2\epsilon+1)}\right) L_n(\hat{y},y)- \left(\frac{\epsilon}{Y^2}\right) \min\limits_{\beta\in[0,1]} \left\{ L_n(\hat{y}_{\beta},y)\right\}  \nn\\
&\leq O(1).
\end{align}
Hence, it follows that
\begin{align}
&L_n(\hat{y},y)-\left( \frac{2 \epsilon+1}{1-z^2}\right) \min\limits_{\beta\in[0,1]}\left\{L_n(\hat{y}_{\beta},y)\right\}\\
&\leq \frac{(2 \epsilon+1)Y^2}{n\epsilon(1-z^2)}O(1) \leq O\left( \frac{1}{\epsilon} \right),
\end{align}\normalsize
which is the desired bound.

Note that using
\begin{align*}
b &= \frac{\epsilon}{Y^2},\;a = \frac{(1-z^2) \epsilon}{Y^2 (2\epsilon+1)},\;s = \left(\frac{Y^2}{2} + \frac{1}{4b}\right),
\end{align*}
we get
\begin{align*}
\mu &= \frac{2+2\sqrt{1-4as}}{s} = \frac{4 \epsilon}{2\epsilon+1}\frac{2+2z}{Y^2}, \label{eq:mue}
\end{align*}\normalsize
after some algebra, as in the statement of the theorem.  This
concludes the proof of the theorem. $\Box$ \\

In the following lemma, we show that the order of the upper bound
using the KL divergence as the distance measure under the same
methodology cannot be improved by presenting an example in which the
bound on $b$ is of the same order as that given in the theorem.\\

\noindent
{\bf Lemma:} For positive real constants $a$, $b$ and $\mu$ which satisfies \eqref{desired_bound} for all $|y_{t}| \leq Y$, $|\hat{y}_{1,t}|\leq Y$ and $|\hat{y}_{2,t}| \leq Y$ and $\lambda_{t} \in [\lp, (1-\lp)]$, we require
\[
b \geq  4a + \frac{a}{4 \lp (1 -\lp)}.
\]\\
\noindent
{\bf Proof:} Since the inequality in \eqref{desired_bound} should be
satisfied for all possible $y_{t}$, $\hat{y}_{1,t}$, $\hat{y}_{2,t}$, $\beta$ and
$\lambda_{t}$, the proper values of $a$, $b$ and $\mu$ should satisfy
\eqref{desired_bound} for any particular selection of $y_{t}$,
$\hat{y}_{1,t}$, $\hat{y}_{2,t}$, $\beta$ and $\lambda_{t}$. First we consider
\begin{align*}
y_{t}&=\hat{y}_{1,t}=Y,\;\hat{y}_{2,t}=0,\;\beta=1,\;\lambda_{t}=\lp,
\end{align*}
(or,similarly, $y_{t}=\hat{y}_{1,t}=Y$, $\hat{y}_{2,t}=-Y$ and
$\lambda_{t}=\lp$). In this case, we have
\begin{align}
&a(Y - \lp Y)^2 \nn\\
&\leq - \ln (\lp + (1- \lp) e^{\mu (Y - \lp Y) \lp (1 - \lp) (-Y)}) \nonumber \\
& \leq - \lp \ln 1 - \mu (1- \lp)^2 \lp Y (1-\lp) (-Y) \label{tbound-1} \\
& = \mu (1- \lp)^3 \lp Y^2, \label{tbound-2}
\end{align}\normalsize
where \eqref{tbound-1} follows from the Jensen's Inequality for concave function $\ln(\cdot)$. By  \eqref{tbound-2}, we have
\begin{align}
\mu \geq \frac{a}{\lp(1- \lp)}. \label{main_bound-1}
\end{align}\normalsize

For another particular case where
\begin{align*}
y_{t}&=-Y/2,\;\hat{y}_{1,t}=0,\;\hat{y}_{2,t}=Y,\;\beta=1,\;\lambda_{t}=1/2,
\end{align*} we have
\begin{align}
&a(- Y)^2 - b (-\frac{Y}{2})^2 \nn\\
& \leq - \ln (\frac{1}{2} + \frac{1}{2} e^{\mu (- Y) \frac{1}{4} (-\frac{Y}{2})}) \nonumber \\
& \leq - \frac{1}{2}\mu \frac{Y^2}{8}, \label{tbound-3}
\end{align}\normalsize
where \eqref{tbound-3} also follows from the Jensen's Inequality. By \eqref{tbound-3}, we have
\begin{align}
b & \geq 4a + \frac{\mu}{4}\geq 4a + \frac{a}{4 \lp (1 -\lp)}, \label{main_bound-2}
\end{align}
\normalsize
where \eqref{main_bound-2} follows from \eqref{main_bound-1}, which finalizes the proof. $\Box$

\section{Simulations} \label{sec:examples}
In this section, we illustrate the performance of the learning algorithm \eqref{eq:1} and the introduced results through examples.
We demonstrate that the upper bound given in \eqref{eq:theorem} is asymptotically tight by providing specific sequences for
the desired signal $y_t$ and the outputs of constituent algorithms $\hat{y}_{1,t}$ and $\hat{y}_{2,t}$. We also demonstrate
that to get a tighter asymptotic bound, we require a smaller learning rate $\mu$, as suggested by our theoretical analysis.

\begin{figure}[t]
\centering
\subfloat[]{\centerline{\epsfxsize=9 cm \epsfbox{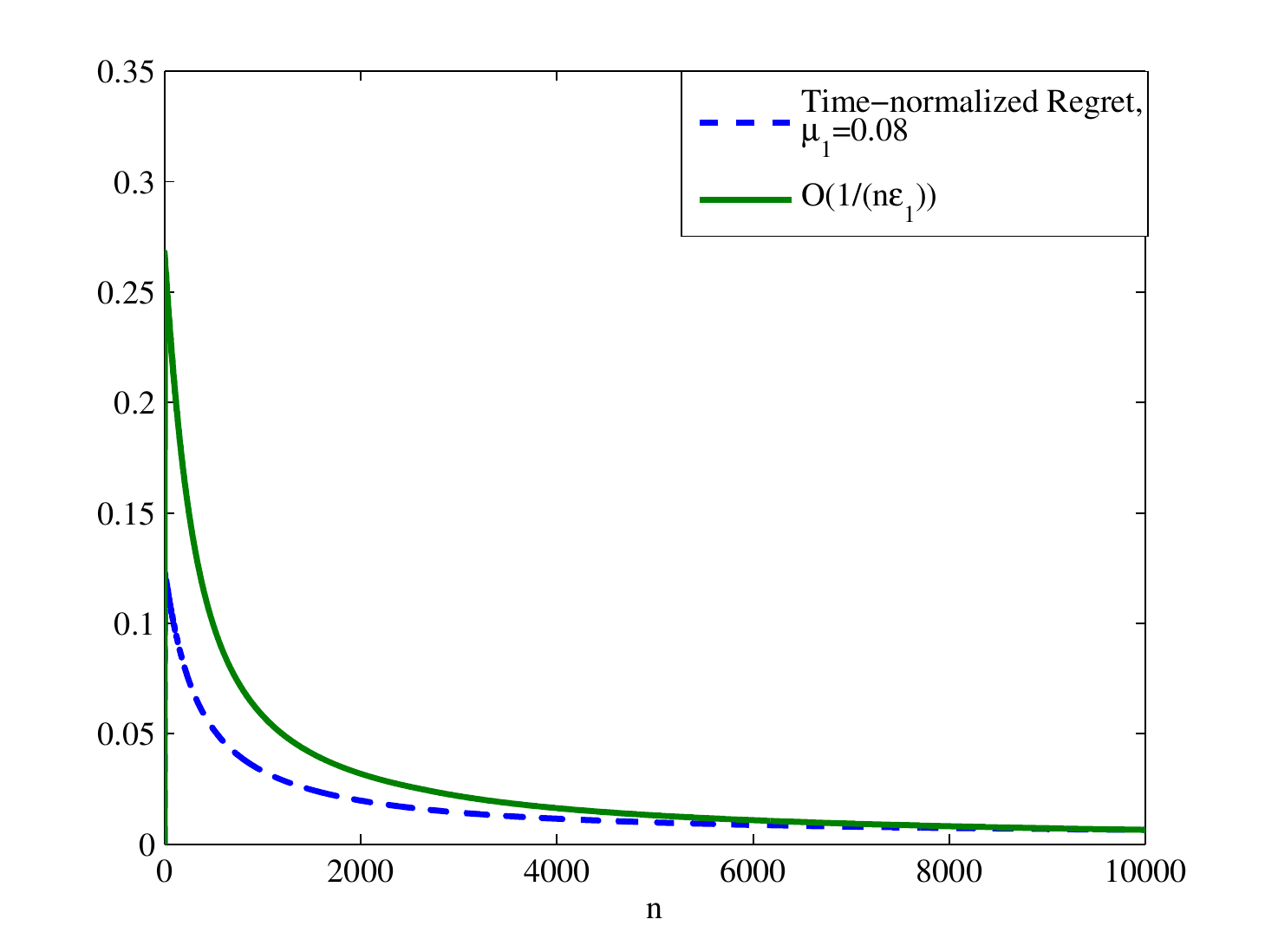}}\label{fig:first}}\\
\subfloat[]{\centerline{\epsfxsize=9 cm \epsfbox{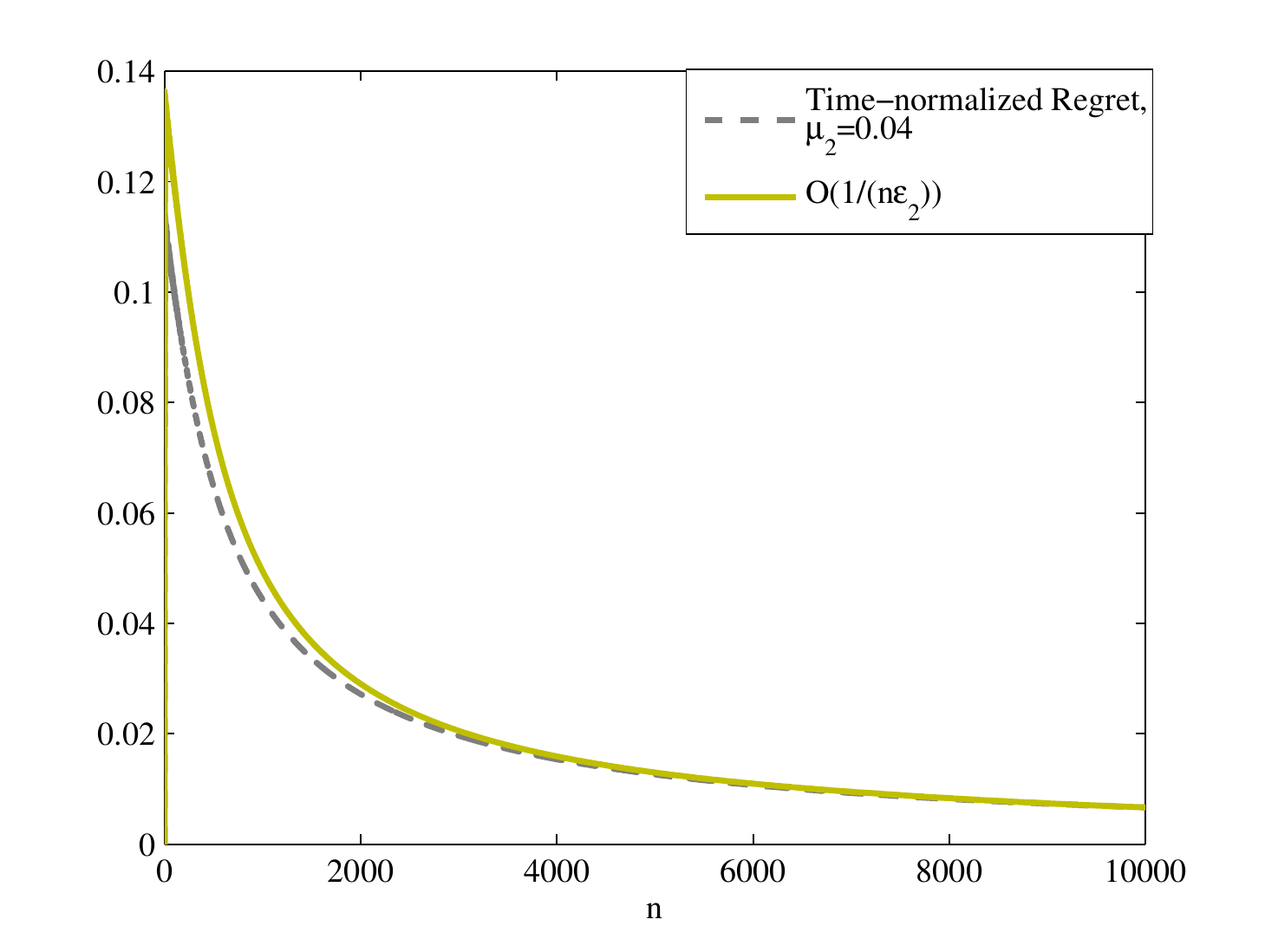}}\label{fig:second}}
\caption{Tightness of the regret bound. (a) $\mu_1=0.08$. (b) $\mu_2=0.04$.}
\label{fig:combinationfigure}
\end{figure}

In the first case, we present the regret of the learning algorithm \eqref{eq:1} defined in \eqref{eq:regret} and the corresponding
upper bound given in \eqref{eq:theorem}. We first set $Y=0.5$, $\lambda^+=0.08$ and $\mu=0.08$. Here, the desired signal is given by
\begin{align*}
y_t=Y
\end{align*}
for $t=1,\ldots,10000$. For this specific example, the parallel running constituent algorithms produce the sequences
\begin{align*}
\hat{y}_{1,t}&=Y,\;\hat{y}_{2,t}=(-1)^tY
\end{align*}
for $t=1,\ldots,10000$. Note that, in this case, the best convex combination weight is $\beta_o=1$ and the cumulative loss of the
best convex combination is 0 since $y_t$ and $\hat{y}_{1,t}$ are identical. In Fig.~\ref{fig:first}, we plot the time-normalized regret of the learning algorithm \eqref{eq:1} ``Time-normalized regret, $\mu_1=0.08$'' and the upper bound given in \eqref{eq:theorem} ``$O(1/(n\eps_1))$''. From Fig.~\ref{fig:first}, we observe that the bound introduced
in \eqref{eq:theorem} is asymptotically tight, i.e., as $n$ gets larger, the gap between the upper bound and the time-normalized regret gets smaller.

In the second case, we set $Y=0.54$, $\lambda^+=0.08$ and $\mu=0.04$. Here, the desired signal is given by
\begin{align*}
y_t=0.5
\end{align*}
for $t=1,\ldots,10000$. For this example, the constituent algorithms produce the sequences
\begin{align*}
\hat{y}_{1,t}&=Y,\;\hat{y}_{2,t}=(-1)^t0.5
\end{align*}
for $t=1,\ldots,10000$. In this case, the best convex combination weight is $\beta_o=0.96$, however, unlike the first case, the cumulative loss of the
best convex combination is nonzero. In Fig.~\ref{fig:second}, we plot the time-normalized regret of the learning algorithm \eqref{eq:1} ``Time-normalized regret, $\mu_2=0.04$'' and the corresponding upper bound given in \eqref{eq:theorem} ``$O(1/(n\eps_2))$'' for this example. We observe from Fig.~\ref{fig:second} that the bound introduced in \eqref{eq:theorem} is asymptotically tight. We also observe that, in this case, the upper bound is tighter compared to the first case since the learning rate, and consequently the parameter $\epsilon$ is smaller, as suggested by our theoretical results.

In this section, we illustrated our theoretical results and the performance of the learning algorithm \eqref{eq:1} through examples.
We observed that the upper bound given in \eqref{eq:theorem} is asymptotically tight by presenting two different examples, i.e., two different cases for
the desired signal $y_t$ and the outputs of constituent algorithms $\hat{y}_{1,t}$ and $\hat{y}_{2,t}$. We also observed
that to get a tighter asymptotic bound, we require a smaller learning rate $\mu$, as suggested by the results introduced in Section~\ref{sec:deterministic_analysis}.

\section{Conclusion}
\label{sec:conclusion}
In this paper, we analyze a learning algorithm \cite{convex}
that adaptively combines outputs of two constituent
algorithms running in parallel to model an unknown desired signal from
the perspective of online learning theory and produce results in an individual
sequence manner such that our results are guaranteed to hold for
any bounded arbitrary signal. We relate the time-accumulated
squared estimation error of this algorithm at any time to the
time-accumulated squared estimation error of the optimal convex
combination of the constituent algorithms that can only be chosen in
hindsight. We refrain from making statistical assumptions on the
underlying signals and our results are guaranteed to hold in an
individual sequence manner. We also demonstrate that the proof
methodology cannot be changed directly to obtain a better bound, in
the convergence rate, on the performance by providing counter
examples. To this end, we provide the transient, steady state
and tracking analysis of this mixture in a deterministic sense without any
assumptions on the underlying signals or without any approximations in
the derivations. We illustrate the introduced results through examples.

\bibliographystyle{IEEEtran}
{
\def\ninept{\def\baselinestretch{0.85}}
\ninept
\bibliography{msaf_references}}

\end{document}